\documentclass{article}
\usepackage{times}
\usepackage{natbib}
\usepackage{textcase,microtype, marvosym}

\usepackage{tikz,pgfplots}
\pgfplotsset{compat=newest}
\pgfplotsset{plot coordinates/math parser=false}

\usepackage{amsmath,amssymb,graphicx,MnSymbol}
\usepackage{todonotes}

\definecolor{lred}{RGB}{200,0,0}
\definecolor{dred}{RGB}{130,0,0} \definecolor{dblu}{RGB}{0,0,130}
\definecolor{dgre}{RGB}{0,130,0} \definecolor{dgra}{RGB}{50,50,50}
\definecolor{mgra}{RGB}{100,100,100}
\definecolor{lgra}{RGB}{220,220,220}
\definecolor{MPG}{RGB}{000,125,122}

\usepackage{booktabs,colonequals}

\usepackage{comment}

\newcommand{\g}{\mid}
\newcommand{\de}{\partial}

\newcommand{\Cov}{\operatorname{cov}}

\renewcommand{\Re}{\mathbb{R}}

\newcommand{\N}{\mathcal{N}}

\newcommand{\Trans}{^{\intercal}}

\newcommand{\Tr}{\operatorname{Tr}}

\renewcommand{\O}{\mathcal{O}}
\newcommand{\GP}{\mathcal{GP}}
\newcommand{\Id}{\mathbb{I}}

\usetikzlibrary{arrows,shapes,plotmarks}

\tikzset{>=stealth'}
\tikzstyle{graphnode} =
   [circle,draw=black,minimum size=22pt,text centered,text
     width=22pt,inner sep=0pt]
\tikzstyle{var}   =[graphnode,fill=white]
\tikzstyle{obs}   =[graphnode,fill=black,text=white]
\tikzstyle{fac}   =[rectangle,draw=black,fill=black!25,minimum size=5pt]
\tikzstyle{facprior} =[rectangle,draw=black,fill=black,text=white,minimum size=5pt]
\tikzstyle{edge}  =[draw=white,double=black,thick,-]
\tikzstyle{prior} =[rectangle, draw=black, fill=black, minimum size=
5pt, inner sep=0pt]
\tikzstyle{dirprior} = [circle, draw=black, fill=black, minimum
size=5pt, inner sep=0pt]

\newif\iffinal 

\iffinal
 
\else
 
\fi

\newcounter{mycomment}

\newcounter{PHcomment}

\usepackage{url}
\usepackage{array}
\usepackage{amsmath,amssymb,amsfonts,textcomp}
\usepackage{epic}
\usepackage[english]{babel}
\usepackage{psfrag}
\usepackage{relsize}
\usepackage{nicefrac}
\usepackage{natbib}   
\usepackage{xspace}

\usepackage{dcolumn}
\newcolumntype{d}[1]{D{.}{.}{#1}}

\newcommand{\acro}[1]{\textsc{#1}}

\newcommand{\gp}{\acro{gp}\xspace}

\renewcommand{\sp}{\ast}
\newcommand{\ud}{\mathrm{d}}

\newcommand{\tra}{\Trans}
\newcommand{\data}{\mathcal{D}}
\newcommand{\sX}{\mathcal{X}}
\newcommand{\sU}{\mathcal{U}}
\newcommand{\p}[2]{p\!\left(#1\middle\vert#2\right)}

\newcommand{\pderiv}[2]{\frac{\partial #1}{\partial #2}}

\newcommand{\deq}{:=}

\newcommand{\psff}[2][]{
  \includegraphics[#1]{#2.pdf}
}

\usepackage{algorithm}
\usepackage{algorithmic}

\usepackage[accepted]{icml2014modified}

\icmltitlerunning{Active Learning of Linear Embeddings for Gaussian Processes}
\begin{document}
\twocolumn[
\icmltitle{Active Learning of Linear Embeddings for Gaussian Processes}

\icmlauthor{Roman Garnett}{rgarnett@uni-bonn.de}
\icmladdress{%
  Department of Computer Science,
  University of Bonn,
  R\"omerstra{\ss}e 164,
  53117 Bonn,
  Germany}
\icmlauthor{Michael A. Osborne}{mosb@robots.ox.ac.uk}
\icmladdress{%
  Department of Engineering Science,
  University of Oxford,
  Parks Road,
  Oxford OX1 3PJ,
  UK}
\icmlauthor{Philipp Hennig}{phennig@tue.mpg.de}
\icmladdress{%
  Department of Empirical Inference,
  Max Planck Institute for Intelligent Systems,
  Spemannstra{\ss}e,
  72076 T\"ubingen,
  Germany}

\vskip 0.3in
]

\begin{abstract}
  We propose an active learning method for discovering low-dimensional
  structure in high-dimensional Gaussian process (\gp) tasks. Such
  problems are increasingly frequent and important, but have hitherto
  presented severe practical difficulties.  We further introduce a
  novel technique for approximately marginalizing \gp hyperparameters,
  yielding marginal predictions robust to hyperparameter
  mis-specification. Our method offers an efficient means of
  performing \gp regression, quadrature, or Bayesian optimization in
  high-dimensional spaces.
\end{abstract}

\section{Introduction}
We propose a method to actively learn, simultaneously, about a
function and a low-dimensional embedding of its input domain. High
dimensionality has stymied the progress of model-based approaches to
many common machine learning tasks. In particular, although Bayesian
nonparametric modeling with Gaussian processes (\gp\/s)
\citep{rasmussen2006gaussian} has become popular for regression,
classification, quadrature \citep{BZHermiteQuadrature}, and global
optimization \citep{brochu2010tutorial}, such approaches remain
intractable for large numbers of input variables (with the exception
of local optimization \citep{HennigKiefel}). An old idea for the
solution to this problem is the exploitation of low-dimensional
structure; the most tractable such case is that of a linear
embedding. Throughout this text, we consider a function
$f(x)\colon\Re^D\to\Re$ of a high-dimensional variable $x\in\Re^D$
(for notational simplicity, $x$ will be assumed to be a row
vector). The assumption is that $f$, in reality, only depends on the
variable $u \deq x R\Trans$, of much lower dimensionality $d\ll D$,
through a linear embedding $R\in\Re^{d\times D}$. We are interested in
an algorithm that simultaneously learns $R$ and $f$, and does so in an
active way. That is, it iteratively selects informative locations
$x_\sp$ in a box-bounded region $\sX\subset \Re^{D}$, and collects
associated observations $y_\sp$ of $f_\sp \deq f(x_\sp)$ corrupted by
i.i.d.\ Gaussian noise: $p(y_\sp\g f_\sp) = \N(y_\sp;f_\sp,\sigma^2)$.

The proposed method comprises three distinct steps (Algorithm
\ref{alg:pseudo}): constructing a probability distribution over
possible embeddings (\emph{learning the embedding} $R$); using this
belief to determine a probability distribution over the function
itself (\emph{learning the function} $f$), and then choosing
evaluation points to best inform these beliefs (\emph{active
  selection}). To learn the embedding, we use a Laplace approximation
on the posterior over $R$ to quantify the uncertainty in the embedding
(Section \ref{sec:embedding}). To learn the function, we develop a
novel approximate means of marginalizing over Gaussian process
hyperparameters (including those parameterizing embeddings), to
provide predictions robust to hyperparameter mis-specification
(Section \ref{sec:marg}). This sub-algorithm is more generally
applicable to many Gaussian process tasks, and to the marginalization
of hyperparameters other than embeddings, and so represents a core
contribution of this paper. Finally, for active selection, we extend
previous work \citep{houlsby2011bayesian} to select evaluations that
maximize the expected reduction in uncertainty about $R$ (Section
\ref{sec:bald}). A simple \textsc{matlab} library (built on
\textsc{gpml}\footnote{\url{http://www.gaussianprocess.org/gpml/code}})
implementing Algorithm \ref{alg:pseudo} and replicating the experiments
in Section \ref{sec:active_experiments} will be released along with
this text.

Estimators for $R$ in wide use include {\sc lasso}
\citep{tibshirani1996regression} and the Dantzig selector
\citep{candes2007dantzig} (which assume $d = 1$). These are passive
methods estimating the linear embedding from a fixed dataset. This
paper develops an algorithm that {\it actively} learns $R$ for the
domain of a Gaussian process. The goal is to use few function
evaluations to intelligently explore and identify $R$. Further,
although the embedding is assumed to be linear, the function $f$
itself will be allowed to be non-linear via the \gp prior.

This problem is related to, but distinct from, dimensionality
reduction \citep{ lawrence2012unifying}, for which active learning has
recently been proposed \citep{iwata2012active}. Dimensionality
reduction is also known as visualization or blind source separation,
and is solved using, e.g., principal component analysis (\acro{pca}),
factor analysis, or latent variable models. As in dimensionality
reduction, we consider the problem of finding a low-dimensional
representation of an input or feature matrix $X\in\Re^{N\times D}$;
unlike dimensionality reduction, we do so given an associated vector
of training outputs or labels $Y\in\Re^N$, containing information
about which inputs are most relevant to a function. The problem of
discovering linear embeddings of \gp\/s was discussed by
\citet{snelson2006variable} for the passive case. Active supervised
learning has been widely investigated \citep{mackay1992information,
  guestrin2005near, houlsby2011bayesian}; our work hierarchically
extends this idea to additionally identify the embedding.  A special
case of our method (the case of a diagonal $R$) is the hitherto
unconsidered problem of {\it active} automatic relevance determination
\citep{mackay1992bayesian, neal1995bayesian, williams1996gaussian}.

Identifying embeddings is relevant for numerous Gaussian process
applications, notably regression, classification, and
optimization. Within Bayesian optimization, much recent work has
focused on high-dimensional problems \citep{hutter2011sequential,
  chen2012joint, carpentier2012bandit, bergstra2012random,
  hutter2009automated}. Recently, \citet{wang2013bayesian} proposed
using randomly generated linear embeddings. In contrast, our active
learning strategy can provide an initialization phase that selects
objective function evaluations so as to best learn low-dimensional
structure. This permits the subsequent optimization of high-dimensional
objectives over only the learned low-dimensional embedding.

\begin{algorithm}[tb]
  \caption{Simultaneous active learning of functions and their linear
    embeddings (pseudocode)}
   \label{alg:pseudo}
\begin{algorithmic}
  \REQUIRE{ $d,D$; kernel $\kappa$, mean function $\mu$; prior $p(R)$}
  \STATE $X\gets\emptyset$; $Y \gets \emptyset$
  \REPEAT
  \STATE $q(R)\gets$ {\sc LaplaceApprox}$\bigl(p(R\g X,Y,\kappa,\mu)\bigr)$\\
  \COMMENT{approximate posterior on embedding $R$}
  \STATE $q(f)\, \gets$ {\sc ApproxMarginal}$\bigl(p(f\g R),q(R)\bigr)$\\
  \COMMENT{approximate marginal on function $f$}
  \STATE $x_\sp\quad\,\gets$ {\sc OptimizeUtility}$\bigl(q(f),q(R)\bigr)$\\
  \COMMENT{find approximate optimal evaluation point $x_*$}
  \STATE $y_\sp\quad\;\gets$ {\sc Observe}$\bigl(f(x_\sp)\bigr)$
  \COMMENT{act}
  \STATE $X\gets [X;x_\sp]$; $Y\gets [Y;y_\sp]$ \COMMENT{store data}
  \UNTIL{budget depleted}
  \RETURN $q(R)$, $q(f)$.
\end{algorithmic}
\end{algorithm}

\section{Linear embeddings of Gaussian processes}
\label{sec:embedding}

\begin{figure}
  \centering
  \begin{tabular}{ll} %
    \psff[height=3cm]{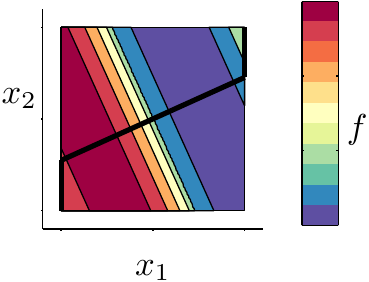}
    &
  \end{tabular}
  \caption{A function $f$ with a one dimensional linear
    embedding (searching over the thick black lines captures all
    variation of the function) and its box-bounded domain $\sX$. Our
    aim is to learn this embedding, represented by embedding matrix
    $R\in \Re^{1\times 2}$, by selecting evaluations of $f$ in some
    search space.
  }
  \label{fig:1d_in_2d}
\end{figure}

In many applications, like image analysis, $D$ can be in the order of
thousands or millions. But even $D=10$ is a high dimensionality for
common Gaussian process models, not only from a computational, but
also from an informational, perspective. Because standard \gp
covariance functions stipulate that function values separated by more
than a few input length scales are negligibly correlated, for high
$D$, almost all of $\sX$ is uncorrelated with observed data. Hence
data is effectively ignored during most predictions, and learning is
impossible. Practical experience shows, however, that many functions
are insensitive to some of their inputs \citep{wang2013bayesian}, thus
have low \emph{effective} dimensionality (Figure
\ref{fig:1d_in_2d}). Our goal is to discover a $R\in \Re^{d\times D}$
such that, for low-dimensional $\sU \subset \Re^{d}$, $u = xR\tra,\
\forall u \in \sU,\ x\in \sX$ and $f(x) = \tilde{f}(u)$ for a new
function, $\tilde{f}\colon \sU \to \Re$.  The discussion here will be
restricted to pre-defined $d$: in reality, this is likely to be
defined as the maximum number of dimensions that can be feasibly
considered in light of computational constraints. If the actual $d$ is
lower than this limit, $R$ can be padded with rows of zeros.

We adopt a \gp prior on $\tilde{f}$ with mean and covariance functions
$\tilde{\mu}$ and $\tilde{\kappa}$, respectively.  The linear
embedding induces another \gp prior $p(f) = \GP(f; \mu, \kappa)$,
where $\mu(x) = \tilde{\mu}(x \,R\tra)$ and $\kappa(x, x') =
\tilde{\kappa}(x\,R\tra, x'\,R\tra)$. For example, if $\tilde{\kappa}$
is the well-known isotropic exponentiated-quadratic
(squared-exponential, radial basis function, Gaussian) covariance,
$\tilde{\kappa}(u,u') \deq \gamma^2 \exp \left[-\frac{1}{2} (u-u')
  (u-u') \Trans \right]$ with output scale $\gamma$, $\kappa$ on $f$
is the Mahalanobis exponentiated-quadratic covariance
\begin{equation}
  \label{eq:3}
  \kappa(x,x') = \gamma^2\exp\left[-\frac{1}{2} (x-x') R\Trans
  R(x-x')\Trans \right].
\end{equation}
If $d=D=1$, then $R\in\Re$ is an inverse length scale. We will return
to this one-dimensional example later to build intuition. A further
special case is a diagonal $R$ (assuming $d=D$), in which case
$\kappa$ is the automatic relevance determination (\acro{ard})
covariance \citep{neal1995bayesian}, widely used to identify the most
important inputs.

Given an appropriate $R$ with acceptably small $d$, learning about $f$
is possible even for large $D$, because the regression problem is
reduced to the manageable space $\Re^d$. This can remain true even in
the case of an uncertain $R$: in particular, assume a prior $p(R)=
\N(R;\hat{R},\Sigma)$. Thus, recalling that $u = xR\Trans$, and using
standard Gaussian identities, if $d=1$, $p(u \g x) =
\N(u;x\hat{R}\Trans,x\Sigma x\Trans)$. If $d>1$, $\Sigma$ is
$\text{Cov}[\text{vect}(R)]$, resulting in another Gaussian for $p(u
\g x)$ that is only slightly more involved than in the $d=1$ case. As
such, regression on $f$ reduces to \gp regression on $\tilde{f}$,
whose domain is the much smaller $\sU \subset \Re^{d}$, but with
uncertain, Gaussian-distributed, inputs. Unlike the work of
\citet{mchutchongaussian}, giving an existing approach to \gp
regression with uncertain inputs, the Gaussian over the inputs here is
correlated; the location of a datum is correlated with all others via
mutual dependence on $R$. And unlike the setting considered by
\citet{girard2005gaussian}, there is no natural ordering of this
domain enabling an iterative procedure. The following section
describes a novel means of regression with uncertain embedding $R$.

\subsection{Approximating the posterior on $R$}
\label{sec:appr-post-r}

The log-likelihood of $R$, after $N$ observations forming a dataset
$\data \deq ({Y},X)\in \Re^N\times \Re^{N\times D}$, is
\begin{align}
  \label{eq:14}
 \log p(Y\g X, R)
 &= \log \N(Y;\mu_X,K_{XX} + \sigma^2 \Id)\\
 &= -\frac{1}{2}\big[ (Y-\mu_X)\Trans(K_{XX}+\sigma^2\Id)^{-1}
 ({Y}-\mu_X)\notag \\
 &\quad + \log\lvert K_{XX}+\sigma^2\Id\rvert + N\log 2\pi\big].\notag
\end{align}
As $\mu_X \deq \mu(X)$ and $K_{XX} \deq \kappa(X, X)$ have non-linear
dependence upon $R$, so does $p(Y\g X,R)$.  Even a simplistic prior on
the elements of $R$ thus gives a complicated posterior. We will use a
Laplace approximation for $p(R \g \data)$ to attain a tractable
algorithm. To construct a Gaussian approximation,
$\N(R;\hat{R},\Sigma)\simeq p(R \g \data)$, we find a mode of the
posterior of $p(R\g \data)$ and set this mode as the mean $\hat{R}$ of
our approximate distribution. The covariance of the Gaussian approximation
is taken as the inverse Hessian of the negative logarithm of the
posterior evaluated at $\hat{R}$,
\begin{equation}
  \label{eq:15}
  \Sigma^{-1} = - \nabla\nabla\Trans \log p(R\g \data)\bigr\rvert_{R=\hat{R}}.
\end{equation}

\subsubsection{Computational Cost}
\label{sec:computational-cost}

How costly is it to construct the Laplace approximation of
Equation~(\ref{eq:15})? Since $D$ may be a large number, active
learning should have low cost in $D$. This section shows that the
required computations can be performed in time linear in $D$, using
standard approximate numerical methods. It is a technical aspect
that readers not interested in details may want to skip over.

Up to normalization, the log posterior is the sum of log prior and log
likelihood \eqref{eq:14}. The former can be chosen very
simplistically; the latter has gradient and Hessian given by, defining $G
\deq \kappa_{XX}+\sigma^2\Id$ and $\Gamma \deq G^{-1}({Y}-\mu_X)$,
\begin{align}
  \notag -2\frac{\de \log p({Y}\g X,R)}{\de \theta} &= -\Gamma\Trans
  \frac{\de \kappa_{XX}}{\de \theta}\Gamma + \Tr \left[G^{-1}
    \frac{\de \kappa_{XX}}{\de
      \theta}\right];\\
  \label{eq:6}
  -2\frac{\de^2 \log p({Y}\g X,R)}{\de \theta\de \eta} &=
  2\Gamma\Trans\frac{\de \kappa_{XX}}{\de \eta}G^{-1} \frac{\de
  \kappa_{XX}}{\de \theta}\Gamma \\
&- \Tr\left[G^{-1}\frac{\de
    \kappa_{XX}}{\de
    \eta}G^{-1} \frac{\de \kappa_{XX}}{\de \theta}\right]\nonumber\\
  &- \Gamma\Trans \frac{\de^2 \kappa_{XX}}{\de \theta\de
  \eta}\Gamma + \Tr\left[G^{-1} \frac{\de^2 \kappa_{XX}}{\de
    \theta\de\eta}\right].\notag
\end{align}
Together with the analogous expressions for a prior $p(R)$, these
expressions can be used to find a maximum of the posterior
distribution (e.g., via a quasi-Newton method), and the Hessian matrix
required for the Laplace approximation to $p(R \g \data)$. The
computational cost of evaluating these expressions depends on the
precise algebraic form of the kernel $\kappa$.

For the exponentiated quadratic kernel of Equation \ref{eq:3},
careful analysis shows that the storage cost for the Hessian of
\eqref{eq:14} is $\O(N^2 dD)$, and its structure allows its
multiplication with a vector in $\O(N^2 dD)$. The corresponding
derivations are tedious and not particularly enlightening. To give an
intuition, consider the most involved term in (\ref{eq:6}): Using the
short-hand $\Delta^{ij} _\ell:=x_{i\ell} - x_{j\ell}$, a straightforward
derivation gives the form
\begin{align*}
  H^1 _{k\ell,ab} :&=-\sum_{ij} \Gamma_i \frac{\de^2
    \kappa(x_i,x_j)}{\de R_{k\ell}\de
    R_{ab}}\Gamma_j\\
  &= \sum_{ijop} R_{ko}\Delta^{ij} _o \Delta^{ij} _\ell \Gamma_i
  \kappa(x_i,x_j)\Gamma_j
  R_{ap}\Delta^{ij} _{p}\Delta_b ^{ij} \\
  &\quad - \sum_{ij} \delta_{ka} \Delta^{ij}_b \Gamma_i
  \kappa(x_i,x_j)\Gamma_j \Delta_{\ell} ^{ij}.
\end{align*}
Multiplication of this term with some vector $g_{ab}$ (resulting from
stacking the elements of the $D\times d$ matrix $g$ into a vector)
requires storage of the $d\times N \times N$ array $R\Delta$ with
elements $(R\Delta)^{ij} _k$, the $D\times N \times N$ array $\Delta$
with elements $\Delta_\ell ^{ij}$, and the $N\times N$ matrix
$\Gamma\Gamma\Trans\otimes K$. Multiplication then takes the form
\begin{align}
  \label{eq:10}
  [H^1 g]_{k\ell} &= \sum_{j=1} ^N \sum_{i=1} ^{N} (R\Delta)_{k} ^{ij}
  \Delta_\ell ^{ij} \Gamma_i \Gamma_j \kappa_{x_ix_j}\\
  &\quad\cdot\underbrace{\left[\sum_{a=1} ^d (R\Delta)^{ij} _{a}\left[\sum_{b =
          1} ^D \Delta_{b}
        ^{ij}g_{ab}\right] \right]}_{\text{compute once in $\O(N^2
      dD)$, store in $\O(N^2)$.}}\notag
\end{align}
Since the $N\times N$ matrix in the square brackets is independent of
$k\ell$, it can be re-used in the $dD$ computations required to
evaluate the full matrix-vector product, so the overall computation
cost of this product is $\O(N^2 dD)$. The other required terms are of
similar form. This means that approximate inversion of the Hessian,
using an iterative solver like the Lanczos or conjugate gradient
methods, is achievable in time linear in $D$. The methods described
here are computationally feasible even for high-dimensional
problems. Our implementation of the active method, which will be
released along with this text, does not yet allow this kind of
scalability, but the derivations above show that it is feasible in
principle.

\section{Approximate marginalization of Gaussian process
  hyperparameters}
\label{sec:marg}

To turn the approximate Gaussian belief on $R$ into an approximate
Gaussian process belief on $f$, the active learning algorithm
(constructed in Section \ref{sec:bald}) requires an (approximate)
means of integrating over the belief on $R$. The elements of $R$ form
hyperparameters of the \gp model. The problem of dealing with
uncertainty in Gaussian process hyperparameters is a general one, also
faced by other, non-active, Gaussian process regression models. This
section presents a novel means of approximately integrating over the
hyperparameters of a \gp. The most widely used approach to learning
\gp hyperparameters is type-II maximum likelihood estimation (evidence
maximization), or maximum {\it a-posteriori} (\acro{map}) estimation,
which both approximate the likelihood as a delta function. However,
ignoring the uncertainty in the hyperparameters in this way can lead
to pathologies \citep{mackay2003information}.

For compact notation, all hyperparameters to be marginalized will be
subsumed into a vector $\theta$. We will denote as
$m_{f|\data,\theta}(x)$ the \gp posterior mean prediction for $f(x)$
conditioned on data $\data$ and $\theta$, and similarly as
$V_{f|\data,\theta}(x)$ the posterior variance $V$ of $f(x)$
conditioned on $\data$ and $\theta$.

We seek an approximation to the intractable posterior for $f_\sp =
f(x_\sp)$, which requires marginalization over $\theta$,
\begin{equation}
  \label{eq:7}
  p(f_\sp\g\data) = \int p(f_\sp\g\data,\theta)\,p(\theta\g\data)\,\ud \theta.
\end{equation}
Assume a Gaussian conditional, $\p{\theta}{\data} = \N(\theta;
\hat{\theta}, \Sigma)$ on the hyperparameters, such as the approximate
distribution over $R$ constructed in the preceding section. To make
the integral in (\ref{eq:7}) tractable, we seek a linear approximation
\begin{align}
  \p{f_\sp}{\data,\theta} &= \N\bigl(f_\sp; m_{f\g\data,\theta}(x_\sp),
  V_{f\g\data,\theta}(x_\sp)\bigr)\\
  &\simeq q(f_\sp; \theta) \deq \N(f_\sp; a\Trans\theta + b, \nu^2),
\end{align}
using free parameters $a,b,\nu^2$ to optimize the fit. The motivation
for this approximation is that it yields a tractable marginal, $
\p{f_\sp}{\data} \simeq \N(f_\sp; a\Trans \hat{\theta} + b, \nu^2 +
a\Trans \Sigma a)$. Further, the posterior for $\theta$ typically has
quite narrow width, over which $\p{f_\sp}{\data,\theta}$'s dependence
on $\theta$ can be reasonably approximated. We choose the variables
$a,b,\nu^2$ by matching a local expansion of $q(f_*\g\theta)$ to
$\p{f_\sp}{\data,\theta}$. The expansion will be performed at
$\theta=\hat{\theta}$, and at a $f_\star = \hat{f}_\star$ to be
determined. Specifically, we match as
\begin{align}
  \left. \pderiv{}{f_\sp} q(f_\sp; \theta)\right\g_{\hat{\theta},
    \hat{f}_\star} &=\left. \pderiv{}{f_\sp}
    \p{f_\sp}{\data,\theta}\right\g_{\hat{\theta}, \hat{f}_\star},
  \label{eq:df}\\
  \left. \pderiv{}{\theta_i} q(f_\sp; \theta)\right\g_{\hat{\theta},
    \hat{f}_\star} &=\left. \pderiv{}{\theta_i}
    \p{f_\sp}{\data,\theta}\right\g_{\hat{\theta}, \hat{f}_\star},
  \label{eq:dth}\\
  \left. \pderiv{^2}{f_\sp^2} q(f_\sp; \theta)\right\g_{\hat{\theta},
    \hat{f}_\star} &=\left. \pderiv{^2}{f_\sp^2}
    \p{f_\sp}{\data,\theta}\right\g_{\hat{\theta}, \hat{f}_\star},
  \label{eq:df2}\\
  \left. \pderiv{^2}{f_\sp \partial \theta_i} q(f_\sp;
    \theta)\right\g_{\hat{\theta}, \hat{f}_\star}
  &=\left. \pderiv{^2}{f_\sp \partial \theta_i}
    \p{f_\sp}{\data,\theta}\right\g_{\hat{\theta}, \hat{f}_\star}\,.
\label{eq:dfdth}
\end{align}
An alternative set of constraints could be constructed by including
second derivatives with respect to $\theta$. But this would require
computation scaling as $O((\#\theta)^2)$, prohibitive for large
numbers of hyperparameters, such as the $D\times d$ required to
parameterize $R$ for large $D$. We define
\begin{equation}
 \hat{m} \deq m_{f|\data,\hat{\theta}}\quad \text{and}\quad
\pderiv{\hat{m}}{\theta_i} \deq
\pderiv{m_{f|\data,\theta}}{\theta_i}\big|_{\theta=\hat{\theta}},
\end{equation}
along with analogous expressions for $\hat{V}$ and
$\pderiv{\hat{V}}{\theta_i}$. Turning to solving for $a,b,\nu^2$ and
$f_\star$, note that, firstly, \eqref{eq:df} implies that $a\Trans
\hat{\theta}+b = \hat{m}$, and that \eqref{eq:df2} implies that $\nu^2
= \hat{V}$. Rearranging \eqref{eq:dth} and \eqref{eq:dfdth},
respectively, we have
\begin{align}
 2 a_i & = \pderiv{\hat{V}}{\theta_i}
 \Biggl( \frac{1}{\hat{f}_\star - \hat{m}} - \frac{\hat{f}_\star - \hat{m}}{\hat{V}} \Biggr)
+ 2 \pderiv{\hat{m}}{\theta_i}\,,
\label{eq:dth_rearrange}\\
 2 a_i & = 2 \pderiv{\hat{V}}{\theta_i}
 \frac{\hat{f}_\star - \hat{m}}{\hat{V}}
+ 2 \pderiv{\hat{m}}{\theta_i}\,.
\label{eq:dfdth_rearrange}
\end{align}
\eqref{eq:dth_rearrange} and \eqref{eq:dfdth_rearrange} can be solved only for
\begin{align}
  a_i & = a_{i\pm} \deq \pm \frac{1}{\sqrt{3 \hat{V}}} \pderiv{\hat{V}}{\theta_i} + \pderiv{\hat{m}}{\theta_i} \\
  f_\sp & =\hat{f}_{\sp \pm}\deq \hat{m}(x_\sp) \pm
  \sqrt{\frac{\hat{V}(x_\sp)}{3}}.
\end{align}
In particular, note that the intuitive choice $f_\sp =
\hat{m}(x_\sp)$, for which $\pderiv{}{f_\sp} \p{f_\sp}{\data,\theta} =
0$, gives $q$ inconsistent constraints related to its variation with
$\theta$. Introducing the separation of
$\surd(\nicefrac{\hat{V}(x_\sp)}{3})$ provides optimal information
about the curvature of $\p{f_\sp}{\data,\theta}$ with $\theta$. Hence
there are two possible values, $\hat{f}_{\sp \pm}$, to expand around,
giving a separate Gaussian approximation for each. We average over the
two solutions, giving an approximation that is a mixture of two
Gaussians. We then further approximate this as a single moment-matched
Gaussian.

\begin{figure*}
  \centering
  \psff[height=4cm, width = 18cm]{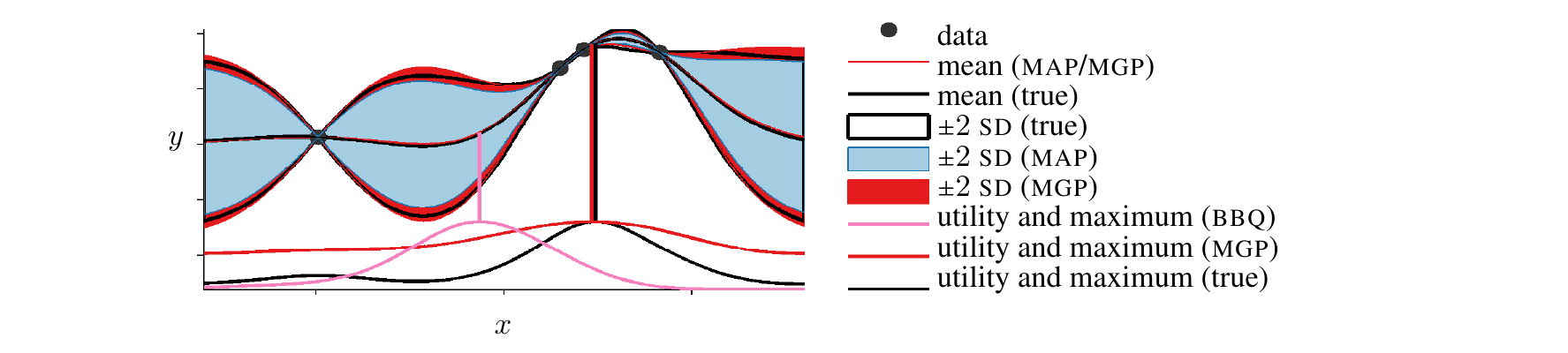}
  \caption{Approximate marginalization (\acro{mgp}) of covariance
    hyperparameters $\theta$ increases the predictive variance to closely match
    the ``true'' posterior (obtained using slice sampling with
    $10^5$ samples). \acro{bbq} \citep{osborne2012active} provides a
    standard deviation differing from the \acro{map} standard
    deviation by less than 3.1\% everywhere, and would hence be
    largely invisible on this plot.  The bottom of the figure displays
    the (normalized) mutual information $I\bigl(\Theta; F(x)\bigr)$
    (equal to the \acro{bald} utility function $\upsilon(x)$) for the
    various methods, and their maxima, giving the optimal positions
    for the next function evaluations. The \acro{mgp} position is very
    close to the true position.}
  \label{fig:marg}
\end{figure*}

The consequence of this approximation is that
\begin{equation}
p(f_\sp \g \data) \simeq
\N\bigl(f_\sp; \tilde{m}_{f|\data}(x_\sp) , \tilde{V}_{f|\data}(x_\sp)\bigr),
\end{equation}
where the marginal mean for $f_\sp$ is $\tilde{m}_{f|\data}(x_\sp)
\deq \hat{m}(x_\sp) $, and the marginal variance is
\begin{align}
  \tilde{V}_{f|\data}(x_\sp) &\deq \frac{4}{3} \hat{V}(x_\sp)
  + \pderiv{\hat{m}(x_\sp)}{\theta}\Trans \Sigma\,
  \pderiv{\hat{m}(x_\sp)}{\theta} \nonumber\\
  &\quad+
  \frac{1}{3\hat{V}(x_\sp)}\pderiv{\hat{V}(x_\sp)}{\theta}\Trans \Sigma\,
  \pderiv{\hat{V}(x_\sp)}{\theta} \,.
  \label{eq:mgp}
\end{align}
Figure \ref{fig:marg} provides an illustration of our approximate
marginal \gp (henceforth abbreviated as \acro{mgp}).

Our approach is similar to that of \citet{osborne2012active}
(\acro{bbq}), for which $\tilde{V}_{f|\data} =
V_{f|\data,\hat{\theta}} +  \pderiv{\hat{m}}{\theta}\Trans \Sigma
\pderiv{\hat{m}}{\theta} $.  However, \acro{bbq} ignores the
variation of the predictive variance with changes in
hyperparameters.

To compare the two methods, we generated (from a \gp) $10 \times D$
random function values, $\data$, where $D$ is the problem
dimension. We then trained a \gp with zero prior mean and \acro{ard}
covariance on that data, and performed prediction for $10 \times D$
test data. Test points, $(x_\sp, y_\sp)$, were generated a small
number (drawn from $\mathcal{U}(1,3)$) of input scales away from a
training point in a uniformly random direction. The \acro{mgp} and
\acro{bbq} were used to approximately marginalize over all \gp
hyperparameters (the output scale and $D$ input scales), computing
posteriors for the test points. We considered $D \in \{5,10,20\}$ and
calculated the mean \acro{skld} over fifty random repetitions of each
experiment. We additionally tested on two real
datasets:\footnote{\url{http://archive.ics.uci.edu/ml/datasets}.}
yacht hydrodynamics \citep{gerritsma1981geometry} and (centered)
concrete compressive strength \citep{yeh1998modeling}. In these two, a
random selection of 50 and 100 points, respectively, was used for
training and the remainder for testing. All else was as above, with
the exception that ten random partitions of each dataset were
considered.

We evaluate performance using the symmetrized Kullback--Leibler
divergence (\acro{skld}) between approximate posteriors and the
``true'' posterior (obtained using a run of slice sampling
\citep{neal2003slice} with $10^5$ samples and $10^4$ burn-in); the
better the approximate marginalization, the smaller this divergence.
We additionally measured the average negative predictive
log-likelihood, $-\mathbb{E}\bigl[\log p(y_\sp \g x_\sp, \data)
  \bigr]$, on the test points $(x_\sp, y_\sp)$. Results are displayed
in Table \ref{tbl:approx_marginal_results}; it can be seen that the
\acro{mgp} provides both superior predictive likelihoods and
posteriors closer to the ``true'' distributions. The only exception is
found on the yacht dataset, where the \acro{mgp}'s \acro{skld} score
was penalized for having predictive variances that were consistently
slightly larger than the ``true'' variances. However, these
conservative variances, in better accommodating test points that were
unexpectedly large or small, led to better likelihoods than the
consistently over-confident \acro{map} and \acro{bbq} predictions.

\begin{table*}[ht!]
\caption{{
    Mean negative log-likelihood for test points and mean \acro{skld}
    (nats)  between approximate and true posteriors. Both metrics were
    averaged over test points, as well as over fifty and ten random
    repeats for synthetic and real experiments, respectively. 
  }}
\label{tbl:approx_marginal_results}
\begin{center}
  \footnotesize
  \begin{tabular}{l r d{7} d{7} d{7} d{7} d{7} d{7}}
  \toprule
  & &
  \multicolumn{3}{c}{$-\mathbb{E}\bigl[\log p(y_\sp \g x_\sp, \data) \bigr]$}
    &
    \multicolumn{3}{c}{\acro{skld}}
    \\
    \cmidrule(rl){3-5}
    \cmidrule(l){6-8}
problem & dim
& \multicolumn{1}{c}{\acro{map}}
& \multicolumn{1}{c}{\acro{bbq}}
& \multicolumn{1}{c}{\acro{mgp}}
& \multicolumn{1}{c}{\acro{map}}
& \multicolumn{1}{c}{\acro{bbq}}
& \multicolumn{1}{c}{\acro{mgp}}
\\ \midrule
synthetic & 5
& 3.58 & 2.67 & {\bf 1}.{\bf 73}
& 0.216 & 0.144 & {\bf 0}.{\bf 0835} \\
synthetic & 10
& 3.57 & 3.10 & {\bf 1}.{\bf 86}
& 0.872 & 0.758 & {\bf 0}.{\bf 465} \\
synthetic & 20
& 1.46 & 1.41 & {\bf 0}.{\bf 782}
& 1.01 & 0.947 & {\bf 0}.{\bf 500} \\
yacht & 6
& 123.0 & 97.8 &  {\bf 56}.{\bf 8}
& 0.0322 & {\bf 0}.{\bf 0133} & 0.0323 \\
concrete & 8
& 2.96\cdot10^9 & 2.96\cdot10^9 & {\bf 1}.{\mathbf{67\cdot10^9}}
& 0.413 & 0.347 & {\bf 0}.{\bf 337}\\
\bottomrule
\end{tabular}
\end{center}
\end{table*}

\section{Active learning of Gaussian process hyperparameters}
\label{sec:bald}

\begin{figure*}[t]
  \centering
  \begin{tabular}{cc}
    \psff[height=2.8cm, width = 7.5cm]{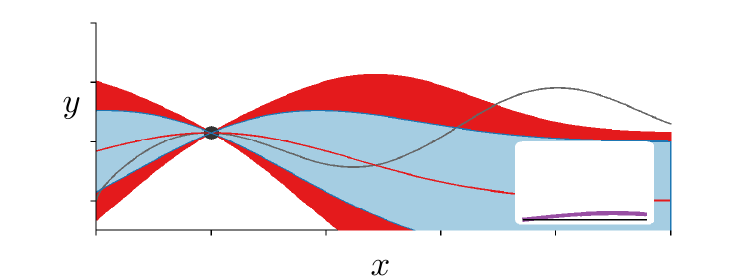} &
    \psff[height=2.8cm, width = 7.5cm]{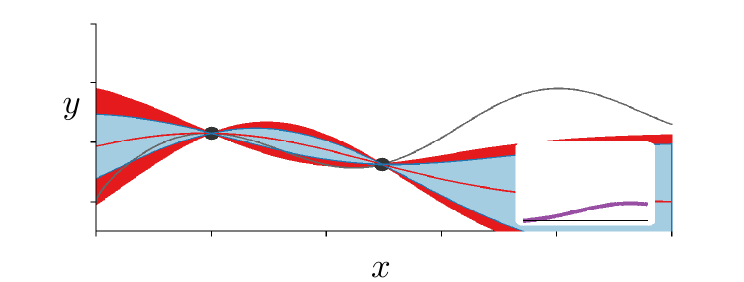}\\
    \psff[height=2.8cm, width = 7.5cm]{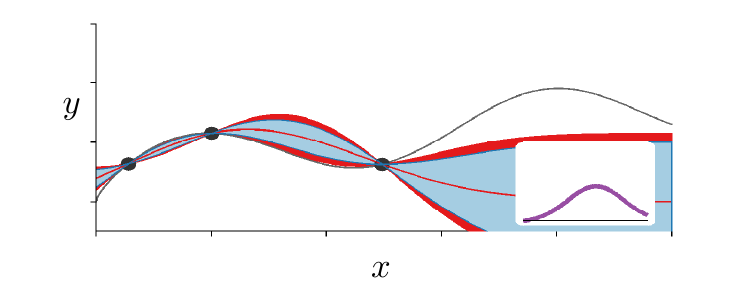} &
    \psff[height=2.8cm, width = 7.5cm]{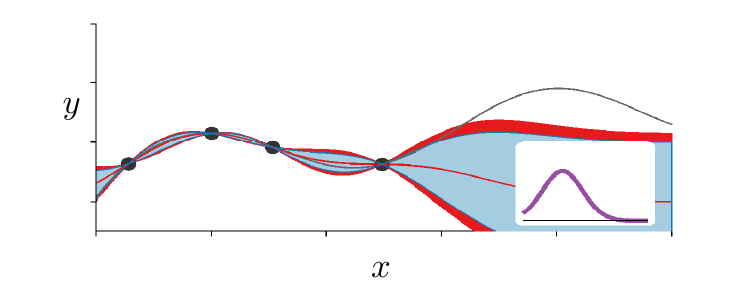}\\
    \psff[height=2.8cm, width = 7.5cm]{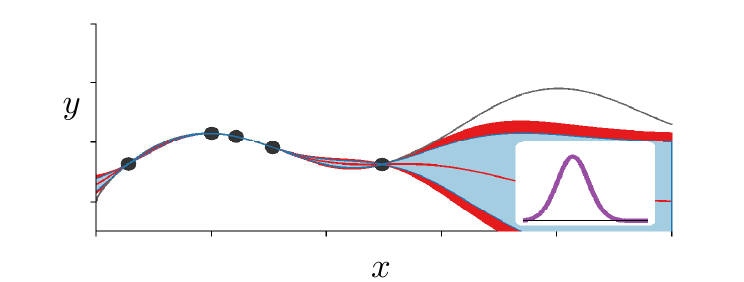} &
    \psff[height=2.8cm, width = 7.5cm]{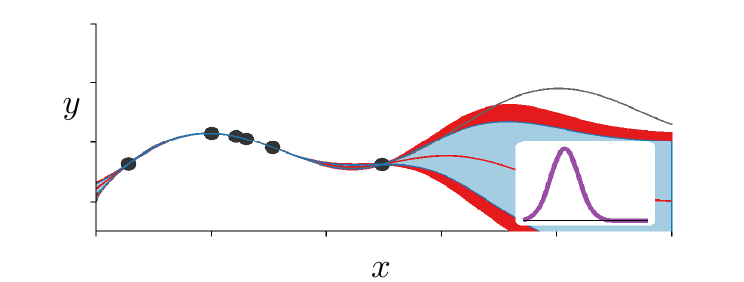}
  \end{tabular}
  \caption{Active learning of the length scale of a one-dimensional
    \gp (beginning at the top left and continuing across and then
    down): the next sample is taken where the \acro{map} and
    approximate variances maximally disagree, normalized by the
    \acro{map} variance. Samples are taken at a variety of separations
    to refine belief about the length scale. The inset plots (all of
    which share axes) display the approximate posteriors over
    log-length scales, which tighten with increasing numbers of
    samples. The legend is identical to that of Figure \ref{fig:marg}.}
  \label{fig:oned_bald}
\end{figure*}

Now we turn to the question of actively selecting observation
locations to hasten our learning of $R$.  We employ an active learning
strategy due to \citet{houlsby2011bayesian}, known as \emph{Bayesian
  active learning by disagreement} (\acro{bald}). The idea is that, in
selecting the location $x$ of a function evaluation $f$ to learn
parameters $\theta$, a sensible utility function is the expected
reduction in the entropy of $\theta$,
\begin{equation}
  \upsilon(x) \deq H(\Theta) - H(\Theta \mid F) = H(F)- H(F \mid
  \Theta),\label{eq:BALD_obj}
\end{equation}
also equal to the mutual information $I(\Theta; F)$ between $f$ and
$\theta$. Mutual information, unlike differential entropies, is
well-defined: the \acro{bald} objective is insensitive to changes in
the representation of $f$ and $\theta$. The right-hand-side of
\eqref{eq:BALD_obj}, the expected reduction in the entropy of $f$
given the provision of $\theta$, is particularly interesting. For our
purposes, $\theta$ will parameterize $R\in\Re^{d\times D}$: that is,
$\theta$ is very high-dimensional, making the computation of
$H(\Theta)$ computationally demanding. In contrast, the calculation of
the entropy of $f \in \Re$ is usually easy or even trivial. The
right-hand side of \eqref{eq:BALD_obj} is particularly straightforward
to evaluate under the approximation of Section \ref{sec:marg}, for
which $p(f\mid \data,\theta)$ and the marginal $p(f\mid \data)$ are
both Gaussian. Further, under this approximation, $p(f\mid \data,\theta)$
has variance $\nu^2 = \hat{V}$ that is independent of $\theta$, hence,
$H(F \mid \Theta) = H(F \mid \Theta = \hat{\theta})$.
We henceforth consider the equivalent but transformed
utility function
\begin{equation}
  \upsilon'(x) = \tilde{V}_{f|\data}(x)\, \Big(
  V_{f|\data,\hat{\theta}}(x) \Big)^{-1}. \label{eq:obj}
\end{equation}
The \acro{mgp} approximation has only a slight influence on this
objective -- Figure \ref{fig:marg} compares it to a full
\acro{mcmc}-derived marginal. With reference to \eqref{eq:mgp},
\eqref{eq:obj} encourages evaluations where the posterior mean and
covariance functions are most sensitive to changes in $\theta$ (Figure
\ref{fig:oned_bald}), normalized by the variance in $f$: such points
are most informative about the hyperparameters.
An alternative to \acro{bald} is found in \emph{uncertainty
  sampling}. Uncertainty sampling selects the location with highest
variance, that is, its objective is simply $H(F)$, the first term in
the \acro{bald} objective. This considers only the variance of a
single point, whereas the \acro{bald} objective rewards points that
assist in the learning of embeddings, thereby reducing the variance
associated with all points. An empirical comparison of our method
against uncertainty sampling follows below.

\subsection{Active learning of linear embeddings for Gaussian
  processes}

\begin{figure*}[t]
  \centering
  \begin{tabular}{ll} %
    \psff[height=4cm]{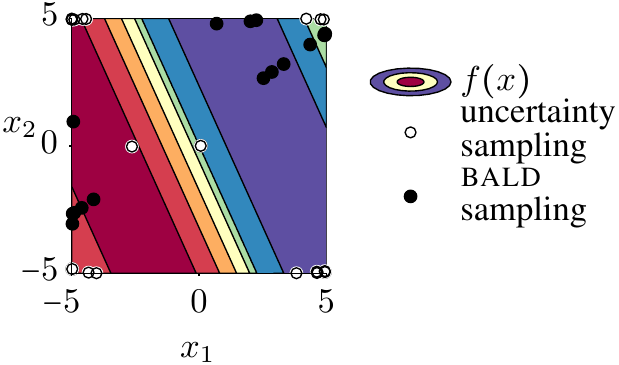}
    &
    \psff[height=3.7cm]{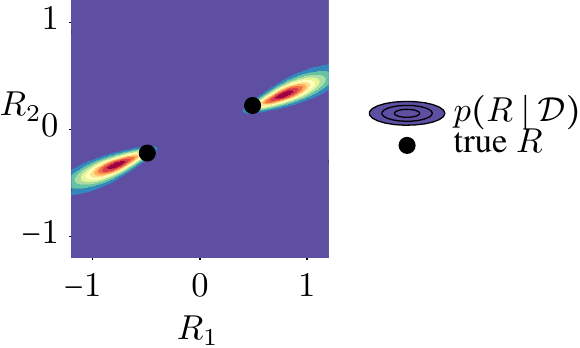}
  \end{tabular}
  \caption{Left: Twenty samples selected by uncertainty sampling and
    \acro{bald} for a function $f$ with a one dimensional linear
    embedding. Note that uncertainty sampling prefers corner
    locations, to maximize the variance in $u$, $x\Sigma x\Trans$, and
    hence the variance in $f$; these points are less useful for
    learning $f$ and its embedding than those selected by
    $\acro{bald}$. Right: the posterior over embeddings returned by
    the $\acro{bald}$ samples, concentrated near the true embedding
    (equivalent under negation). }
  \label{fig:1d_in_2d_bald}
\end{figure*}

To apply \acro{bald} to learning the linear embedding of a Gaussian
process, we consider the case $R \subset\theta$; the \gp
hyperparameters define the embedding described in Section
\ref{sec:embedding}. Figure \ref{fig:1d_in_2d_bald} demonstrates an
example of active learning for the embedding of a two-dimensional
function.

The latent model of lower dimension renders optimizing an objective
with domain $\sX$ (e.g., $f(x)$, or the \acro{bald} objective)
feasible even for high dimensional $\sX$. Instead of direct search
over $\sX$, one can choose a $u \in \sU$, requiring search over only
the low-dimensional $\sU$, and then evaluate the objective at an $x
\in \sX$ for which $u = x\, R\Trans$. A natural choice is the $x$
which is most likely to actually map to $u$ under $R$, that is, the
$x$ for which $p(u \g x)$ is as tight as possible.  For example, we
could minimize $\log \det \Cov[u \g x]$, subject to $\mathbb{E}[u \g
  x] = x\,\hat{R}\Trans$, by solving the appropriate program.  For
$d=1$, this is a quadratic program that minimizes the variance
$x\Sigma x\Trans$ under the equality constraint. Finally, we evaluate
the objective at the solution.

For simplicity, we will henceforth assume $\sX = [-1, 1]^D$. For any
box-bounded problem, there is an invertible affine transformation
mapping the box to this $\sX$; this then requires only that $R$ is
composed with this transformation. Further, define the signature of
the $i$th row of $R$ to be $[\text{sign}(R_{i1}), \text{sign}(R_{i2}),
  \ldots]$. Then, for the $i$th coordinate, the maximum and minimum
value obtained by mapping the corners of $\sX$ through $R$ are
achieved by the corner matching this signature and its negative. This
procedure defines the extreme corners of the search volume $\sU$.

Consider the typical case in which we take $\tilde{\mu}$ as constant
and $\tilde{\kappa}$ as isotropic (e.g., the exponentiated quadratic
\eqref{eq:3}). Since $p(f \g X, R)$ is then invariant to orthogonal
transformations of $R$ in $\Re^d$, there is no unique embedding. In
the special case $d = 1$, $R$ and $-R$ are equivalent.  For most means
and covariances there will be similar symmetries, and likely ever more
of them as $d$ increases. We therefore evaluate the performance of our
algorithms not by comparing estimated to true $R$s, which is difficult
due to these symmetries, but rather in the direct predictive
performance for $f$.

\subsection{Active learning of linear embeddings experiments}
\label{sec:active_experiments}

We now present the results of applying our proposed method for
learning linear embeddings on both real and synthetic data with
dimension up to $D = 318$. Given a function $f\colon \sX \to \Re$ with
a known or suspected low-dimensional embedding, we compare the
following methods for sequentially selecting $N = 100$ observations
from the domain $[-1, 1]^D$: random sampling (\textsc{rand}), a Latin
hypercube design (\textsc{lh}), uncertainty sampling
(\textsc{unc}), and \textsc{bald}. \textsc{unc} and
\textsc{bald} use identical models (Laplace approximation on $R$
followed by \textsc{mgp}) and hyperparameter priors. We also compare
with \textsc{lasso}, choosing the regularization parameter by
minimizing squared loss on the training data. The functions that these
methods are compared on are: \begin{itemize}
  \item Synthetic in-model data drawn from a \gp matching our model
    with an embedding drawn from our prior, for $d \in \lbrace 2, 3
    \rbrace$ and $D \in \lbrace 10, 20 \rbrace$.
  \item The Branin function, a popular test function for global
    optimization ($d = 2$), embedded in $D \in \lbrace 10, 20 \rbrace$
    via an embedding drawn from our prior.
  \item The temperature data\footnote{This dataset available at
    \url{http://theoval.cmp.uea.ac.uk/~gcc/competition}.}  described
    in \citet{snelson2006variable} ($D = 106$), with $d = 2$. The
    associated prediction problem concerns future temperature at a
    weather station, given the output of a circulation model. The
    training and validation points were combined to form the dataset,
    comprising 10\,675 points.
  \item The normalized ``communities and crime'' (\textsc{c\&c})
    dataset from the \textsc{uci} Machine Learning
    Repository\footnote{This dataset available at \url{http://archive.ics.uci.edu/ml/datasets/Communities+and+Crime}.}
    ($D = 96$), with $d = 2$. The task here is to predict
    the number of violent crimes per capita in a set of \textsc{us}
    communities given historical data from the \textsc{us} Census and
    \textsc{fbi}. The \textsc{lemas} survey features were discarded
    due to missing values, as was a single record missing the
    ``AsianPerCap'' attribute, leaving $1\,993$ points.
  \item The ``relative location of \textsc{ct} slices on axial axis''
    dataset from the \textsc{uci} Machine Learning
    Repository\footnote{This dataset available at
      \url{http://archive.ics.uci.edu/ml/datasets/Relative+location+of+CT+slices+on+axial+axis}.}
    ($D = 318$), with $d = 2$. The task is to use features extracted
    from slices of a \textsc{ct} scan to predict its vertical location
    in the human body. Missing features were replaced with zeros.
    Only axial locations in the range $[50, 60]$ were used.  Features
    that did not vary over these points were discarded, leaving
    $3\,071$ points.
\end{itemize}
The \textsc{ct} slices and communities and crime datasets are,
respectively, the highest- and third-highest-dimensional regression
datasets available in the \textsc{uci} Machine Learning Repository
with real attributes; in second place is an unnormalized version of
the \textsc{c\&c} dataset.

\begin{table*}
  \centering
  \caption{The combination of \acro{mgp} and \acro{bald} actively
    learns embeddings whose predictive performance improves on
    alternatives. Average negative log predictive probability and
    average \acro{rmse} on test functions for various $D$ and $d$.\\}
\label{tbl:results}
  \footnotesize
  \begin{tabular}{@{}lr@{/}ld{3}d{3}d{3}d{3}d{3}d{3}d{4}d{4}d{3}@{}}
    \toprule
    \multicolumn{2}{c}{}
    &
    \multicolumn{5}{c}{$-\mathbb{E}\bigl[\log p(y_\sp \g x_\sp, \hat{R}) \bigr]$}
    &
    \multicolumn{5}{c}{\textsc{rmse}}
    \\
    \cmidrule(lr){4-7}
    \cmidrule(l){8-12}
    dataset & $D$ & $d$ & \multicolumn{1}{c}{\textsc{rand}} & \multicolumn{1}{c}{\textsc{lh}} & \multicolumn{1}{c}{\textsc{unc}} & \multicolumn{1}{c}{\textsc{bald}} &
    \multicolumn{1}{c}{\textsc{rand}} & \multicolumn{1}{c}{\textsc{lh}} & \multicolumn{1}{c}{\textsc{unc}} & \multicolumn{1}{c}{\textsc{bald}} & \multicolumn{1}{c}{\textsc{lasso}} \\
    \midrule
    synthetic & 10 & 2 &
    0.272 & 0.224 & -0.564 & {\bf -0}.{\bf 649} &
    0.412 & 0.371 &  0.146 & {\bf  0}.{\bf 138} & 0.842 \\
    synthetic & 10 & 3 &
    0.711 & 0.999 & 0.662 & {\bf 0}.{\bf 465} &
    0.553 & 0.687 & 0.557 & {\bf 0}.{\bf 523} & 0.864 \\
    synthetic & 20 & 2 &
    0.804 & 0.745 & 0.749 & {\bf 0}.{\bf 470} &
    0.578 & 0.549 & 0.551 & {\bf 0}.{\bf 464} & 0.853 \\
    synthetic & 20 & 3 &
    1.07  & 1.10  & 1.04  & {\bf 0}.{\bf 888} &
    0.714 & 0.740 & 0.700 & {\bf 0}.{\bf 617} & 0.883 \\
    Branin & 10 & 2 &
     3.87 &  3.90 & 1.58 & {\bf 0}.{\bf 0165} &
    18.2  & 17.8  & 3.63 & {\bf 2}.{\bf 29}   & 40.0 \\
    Branin & 20 & 2 &
     4.00 &  3.70 & {\bf  3}.{\bf 55} &  3.63 &
    18.3  & 14.8  & {\bf 13}.{\bf 4}  & 15.0  & 39.1 \\
    communities \& crime & 96 & 2 &
    1.09  & \multicolumn{1}{c}{---} & 1.17  & {\bf 1}.{\bf 01}  &
    0.720 & \multicolumn{1}{c}{---} & 0.782 & {\bf 0}.{\bf 661} & 1.16 \\
    temperature & 106 & 2 &
    0.566 & \multicolumn{1}{c}{---} & 0.583 & {\bf 0}.{\bf 318} &
    0.423 & \multicolumn{1}{c}{---} & 0.427 & {\bf 0}.{\bf 328} & 0.430 \\
    \textsc{ct} slices & 318 & 2 &
    1.30  & \multicolumn{1}{c}{---} & 1.26  & {\bf 1}.{\bf 16}  &
    0.878 & \multicolumn{1}{c}{---} & 0.845 & {\bf 0}.{\bf 767} & 0.900 \\
    \bottomrule
  \end{tabular}
\end{table*}

For the synthetic and Branin problems, where the true embedding $R$
was chosen explicitly, we report averages over five separate
experiments differing only in the choice of $R$.  On these datasets,
the \textsc{unc} and \textsc{bald} methods selected points by
successively maximizing their respective objectives on a set of
$20\,000$ fixed points in the input domain, $10\,000$ selected
uniformly in $[-1, 1]^D$ and $10\,000$ selected uniformly in the unit
$D$-sphere.  For a given $D$, these points were fixed across methods
and experimental runs.  This choice allows us to compare methods based
only on their objectives and not the means of optimizing them.

For the real datasets (temperature, communities and crimes, and
\textsc{ct} slices), each method selected from the available points;
\acro{lh} is incapable of doing so and so is not considered on these
datasets.  The real datasets were further processed by transforming
all features to the box $[-1, 1]^D$ via the ``subtract min, divide by
max'' map
and normalizing the outputs to have zero mean and unit variance. For
the synthetic problems, we added i.i.d.\ Gaussian observation noise
with variance $\sigma^2 = (0.1)^2$. For the remaining problems, the
datapoints were used directly (assuming that these real measurements
already reflect noise).

After each method selected 100 observations, we compare the quality of
the learned embeddings by fixing the hyperparameters of a \gp to the
\textsc{map} embedding at termination and measuring predictive
performance.  This is intended to emulate a fixed-budget embedding
learning phase followed by an experiment using only the most likely
$R$. We chose $N = 100$ training points and $1\,000$ test points
uniformly at random from those available; these points are common to
all methods. We report root-mean-square-error and the average negative
predictive log-likelihood on the test points.  The \textsc{rmse}
measures predictive accuracy, whereas the log-likelihood additionally
captures the accuracy of variance estimates.  This procedure was
repeated 10 times for each experiment; the reported numbers are
averages.

The embedding prior $p(R)$ was set to be i.i.d.\ zero-mean Gaussian
with standard deviation $\nicefrac{5}{4}D^{-1}$. This choice roughly
implies that we expect $[-1, 1]^D$ to map approximately within $[-2.5,
  2.5]^d$, a box five length scales on each side, under the unknown
embedding. This prior is extremely diffuse and does not encode any
structure of $R$ beyond prefering low-magnitude values.  At each step,
the mode of the log posterior over $R$ was found using using
\acro{l-bfgs}, starting from both the previous best point and one
random restart drawn from $p(R)$.

The results are displayed in Table \ref{tbl:results}. The active
algorithm achieves the most accurate predictions on all but one
problem, including each of the real datasets, according to both
metrics.  These results strongly suggest an advantage for actively
learning linear embeddings.

\section{Conclusions}

Active learning in regression tasks should include hyperparameters, in
addition to the function model itself. Here we studied simultaneous
active learning of the function and a low-dimensional linear embedding
of its input domain. We also developed a novel means of approximately
integrating over the hyperparameters of a \gp model. The resulting
algorithm addresses needs in a number of domains, including Bayesian
optimization, Bayesian quadrature, and also the underlying idea of
nonparametric Gaussian regression itself. Empirical evaluation
demonstrates the efficacy of the resulting algorithm on both synthetic
and real problems in up to $318$ dimensions, and an analysis of
computational cost shows that the algorithm can, at least in
principle, be scaled to problems of much larger dimensionality as
well.

{\small
  \bibliography{active_dim_redn_initials}
  \bibliographystyle{icml2014}
}

\end{document}